\documentclass[sigconf]{acmart}

\usepackage{latexsym}
\usepackage{graphicx}%
\usepackage{subcaption}
\usepackage{multirow}%
\usepackage{amsmath,amsfonts}%
\usepackage{amsthm}%
\usepackage{mathrsfs}%
\usepackage[title]{appendix}%
\usepackage{xcolor}%
\usepackage{textcomp}%
\usepackage{manyfoot}%
\usepackage{booktabs}%
\usepackage{algorithm}%
\usepackage{algorithmicx}%
\usepackage{algpseudocode}%
\usepackage{listings}%
\usepackage{xspace}
\usepackage{float}
\usepackage{setspace}
\usepackage{tabularx}
\usepackage{array}

\AtBeginDocument{%
  }

\setcopyright{acmlicensed}
\copyrightyear{2018}
\acmYear{2024}
\acmDOI{XXXXXXX.XXXXXXX}

\acmConference[Conference acronym 'XX]{Make sure to enter the correct
  conference title from your rights confirmation emai}{June 03--05,
  2018}{Woodstock, NY}
\acmISBN{978-1-4503-XXXX-X/18/06}

\begin{document}

\title{Low-Dimensional Federated Knowledge Graph Embedding via Knowledge Distillation}

\author{Xiaoxiong Zhang}
\email{zhan0552@e.ntu.edu.sg}
\affiliation{%
  \institution{Nanyang Technological University}
  \country{Singapore}
}

\author{Zhiwei Zeng}
\email{zhiwei.zeng@ntu.edu.sg}
\affiliation{%
  \institution{Nanyang Technological University}
  \country{Singapore}}

\author{Xin Zhou}
\email{xin.zhou@ntu.edu.sg}
\affiliation{%
  \institution{Nanyang Technological University}
  \country{Singapore}}

\author{Chunyan Miao}
\email{ASCYMiao@ntu.edu.sg}
\affiliation{%
  \institution{Nanyang Technological University}
  \country{Singapore}}







\renewcommand{\shortauthors}{Trovato et al.}

\begin{abstract}
Federated Knowledge Graph Embedding (FKGE) aims to facilitate collaborative learning of entity and relation embeddings from distributed Knowledge Graphs (KGs) across multiple clients, while preserving data privacy. Training FKGE models with higher dimensions is typically favored due to their potential for achieving superior performance. However, high-dimensional embeddings present significant challenges in terms of storage resource and inference speed.
Unlike traditional KG embedding methods, FKGE involves multiple client-server communication rounds, where communication efficiency is critical. Existing embedding compression methods for traditional KGs may not be directly applicable to FKGE as they often require multiple model trainings which potentially incur substantial communication costs. In this paper, we propose a lightweight component based on Knowledge Distillation (KD) which is titled \textbf{FedKD} and tailored specifically for FKGE methods. During client-side local training, \textbf{FedKD} facilitates the low-dimensional student model to mimic the score distribution of triples from the high-dimensional teacher model using KL divergence loss. Unlike traditional KD way,  \textbf{FedKD} adaptively learns a temperature to scale the score of positive triples and separately adjusts the scores of corresponding negative triples using a predefined temperature, thereby mitigating teacher over-confidence issue. Furthermore, we dynamically adjust the weight of KD loss to optimize the training process. Extensive experiments on three datasets support the effectiveness of \textbf{FedKD}. 

\end{abstract}





\maketitle
\vspace{-1em}
\section{Introduction}

A knowledge graph (KG) describes real-world facts in the form of triples (head entity, relation, tail entity). Knowledge graph embedding (KGE) aims to encode entities and relations in the KG into continuous vector representations which capture the semantic meanings and relationships inherent in the graph structure, enabling various downstream tasks such as disease diagnosis \cite{sang2018sematyp,abdelaziz2017large}, recommendation system \cite{zhang2016collaborative, xian2019reinforcement,wang2022multi}, question answering system \cite{hao-etal-2017-end} and so on. 

With the promulgation of General Data Protection Regulation (GDPR) \cite{regulation2016679}, KGs from multiple sources are no longer stored centrally on one device as a whole KG, but instead, in a more decentralized manner on multiple clients. Formally, these distributed multi-source KGs are referred to as federated knowledge graphs (FKG). A major line of research in FKG is federated learning-based Federated Knowledge Graph Embedding (FKGE) \cite{fede,fedec,fedkg,fedr}. Through federated learning \cite{tan2022towards, shamsian2021personalized, briggs2020federated, zhang2021survey, zhang2024beyond}, each client obtains useful information from other clients. The aggregated information, together with local triples, is used to update the entities and relations embeddings. The final learned entities and relations embeddings are used to predict non-existent triples for each client. 

To enhance the performance of learned embeddings for FKG, it is common practice to increase embedding dimension during the training phase. However, pre-training high-dimensional embeddings is often impractical in many real-world downstream scenarios, and the low-dimensional embeddings are necessary \cite{zhu2022dualde}. In some cases, the storage capacity and computational resources are limited (such as deploying learned embeddings from FKG on edge computing devices or mobile platforms) while fast inference speed is required (such as online financial systems demanding fast market decisions). Besides, some cases need to further fine-tune the pre-trained embedding of FKG rather than using them directly to address downstream tasks (such as recommendation system) effectively, also expecting to keep the cost of fine-tuning as low as possible. Hence, it is necessary for FKGE to balance the need for low dimension and high performance.


Recently, there are several knowledge-distillation-based methods \cite{wang2021mulde, zhu2022dualde, liu2023iterde} to compress high-dimensional embeddings to low-dimensional ones for KG while there is little research about embedding compression specifically for FKG. The embedding learning process for FKGE is distinct from that of traditional KGE. For FKGE, the process typically entails multiple rounds of communication between clients and a central server. Due to constraints such as limited network bandwidth and data usage on clients, minimizing the transmission of parameters during FKGE training is crucial. Since there is no concern about communication efficiency in traditional KGE, existing KGE embedding compression techniques are often complex and involve multiple model trainings. For example, the method in \cite{wang2021mulde} trains multiple high-dimensional teacher models and use them to teach the low-dimensional student model collaboratively. The approach described in \cite{liu2023iterde} employs iterative knowledge distillation to progressively decrease the model sizes to a desired target, which aims to mitigate the adverse effects of substantial dimension gap between the student and the teacher model on the performance of knowledge distillation. During the process, multiple models of intermediate sizes are trained. As a result, these methods adopted in KGE usually are unsuitable for direct application in FKGE due to the significant communication costs they potentially entail. Therefore, it is necessary to develop more efficient and lightweight embedding compression methods tailored specifically for FKGE.

In this paper, we propose a lightweight method titled Low-Dimensional \textbf{Fed}erated Knowledge Graph Embedding via \textbf{K}nowledge \textbf{D}istillation (\textbf{FedKD}). It, acting as an adjunct to current FKGE methods, enables the transfer of insights from a pre-trained high-dimensional teacher model to a student model during the clients' local training phrase at each communication round. Specifically, during the client's local training phase in each communication round, in addition to applying the original local training loss (i.e., \textbf{hard label loss}), \textbf{FedKD} enables the low-dimensional student model to emulate the score distribution of triples derived from the pre-trained high-dimensional teacher model by employing KL divergence loss. The KL divergence loss is denoted as \textbf{soft label loss}, formally. However, the over-confidence in positive triples by the teacher model can usually lead to a less discriminative distribution of corresponding negative samples and thereby impacting the efficacy of knowledge distillation. To mitigate the issue, in contrast to traditional knowledge distillation methods using KL divergence, we introduce the Adaptive Asymmetric Temperature Scaling mechanism which adaptively learns a temperature to scale the score of the positive triple and separately scales the scores of corresponding negative triples using a predefined temperature. Moreover, the discrepancy in optimization objectives between hard label loss and soft label loss can inadvertently affect training efficiency. To address this, we dynamically adjust the weight of the soft label loss by prioritizing the hard label loss at the early stage of training and progressively transit towards placing greater emphasis on the soft label loss as the training progresses. Extensive experimental results substantiate the effectiveness of \textbf{FedKD}.

Our contributions are summarized as follows:

\begin{itemize}
    \item We introduce a lightweight component, \textbf{FedKD}, designed for existing FKGE methods, which aims to reduce embedding size without substantial compromise to embedding performance. To the best of our knowledge, this is the first method tailored specifically for embedding compression in FKGE.
    \item We propose an Adaptive Asymmetric Temperature Scaling mechanism aimed at mitigating the adverse impact of teacher model over-confidence in positive samples on the performance of on knowledge distillation. Besides,we employ dynamic weight adjustment to alleviate the optimization disparity between the hard and soft label loss.    
    \item We assess the efficacy of \textbf{FedKD} across three datasets through comprehensive experiments. The findings indicate a notable reduction in the embedding dimension of FKGE with only marginal (even no) decrease in performance.

\end{itemize}

\vspace{-1em}
\section{Related Work}

\subsection{Federated Knowledge Graph Embedding}

Federated Learning (FL) represents a novel distributed paradigm in machine learning that enables multiple clients to collaboratively train high-performing models, simultaneously preserving data privacy and security \cite{li2020federated,li2021survey,t2020personalized, sfl, zhang2024privfr}. Based on the FL paradigm, federated knowledge graph embedding aims to facilitate collaborative learning of entity and relation embeddings from multi-source KGs distributed across multiple clients, while ensuring raw triplets are kept confidential and not shared among participants.

\textbf{FedE} \cite{fede}, inspired by \textbf{FedAvg} \cite{mcmahan2017communication}, serves as a pioneering model where a server coordinates clients' local embedding learning by iteratively aggregating their entity embeddings and distributing the aggregated result for the subsequent round of local training. Inspired by \textbf{Moon} \cite{moon}, \textbf{FedEC} introduces embedding contrastive learning into \textbf{FedE} to further improve embedding performance. However, both \textbf{FedE} and \textbf{FedEC} generate a single set of global consensus embeddings for all clients, which may not fully accommodate data heterogeneity. As a result, the embeddings trained to fit an ``average KG'' might not generalize well to KGs that exhibit high heterogeneity with the global training dataset. To address it, several personalized FKGE methods are proposed. \textbf{PFedEG} proposed the personalized embedding aggregation mechanism on the server based on a client-wise relation graph \cite{zhang2025personalized}. \textbf{FedLu} introduces mutual knowledge distillation to facilitate the transfer of information from local entity embeddings to global entity embeddings, followed by the incorporation of knowledge gleaned from the global entity embeddings \cite{zhu2023heterogeneous}. In contrast to the aforementioned methods that predominantly handle scenarios where clients possess partially shared entities but mutually exclusive relations, \textbf{FedR} \cite{fedr} specifically addresses cases where clients not only share entities but also share relations. In this framework, all clients receive identical embeddings of shared relations from the server and subsequently engage in local embedding learning using their respective local triples together with the shared relation embeddings.

Unlike the aforementioned methods that rely on a client-server architecture, \textbf{FKGE} \cite{fedkg} facilitates collaborative embedding learning among clients in a peer-to-peer manner. Drawing inspiration from \textbf{MUSE} \cite{lampleword}, \textbf{FKGE} adopts a Generative Adversarial Network (GAN) framework \cite{goodfellow2020generative} to harmonize the embeddings of shared entities and relations within a pair of knowledge graphs.
Besides, \textbf{FedS} \cite{zhang2025communication} focus on reducing the communication cost of FKGE methods based on the client-server architecture, by introducing entity-wise embedding sparsification strategy.

In this study, our focus centers on methods that utilize a client-server architecture rather than a peer-to-peer architecture.
\vspace{-1em}
\subsection{Knowledge Graph Embedding Compression with Knowledge Distillation}

Knowledge distillation (KD) is a prominent technique employed in model compression \cite{yang2023knowledge, zhu2021student, guo2023boosting}. In this approach, a larger, pre-trained model, referred to as the teacher model, provides guidance to a smaller model, known as the student model. The objective is for the student model to approximate the outputs of the teacher model, thereby achieving similar performance \cite{hinton2015distilling}. 

Many embedding methods demonstrate enhanced performance with higher embedding dimensions. However, the trade-off includes longer inference times and larger model sizes for high-dimensional Knowledge Graph Embedding. Recently, several knowledge distillation methods \cite{wang2021mulde, zhu2022dualde, liu2023iterde} have been proposed to compress high-dimensional embeddings into lower-dimensional representations for knowledge graphs. \textbf{MulDE}\cite{wang2021mulde} employs multi-teacher distillation techniques aimed at improving the efficacy of low-dimensional models. However, it does not adequately address the dimension gap issue, which can significantly affect the effectiveness of the distillation process. \textbf{DualDE} \cite{zhu2022dualde} presents an innovative two-stage distillation approach where only a high-dimensional teacher model is pre-trained. Initially, the teacher model remains fixed while only the student model is updated under its guidance during the first stage. Subsequently, in the second stage, both the teacher and student models reciprocally guide each other's embedding updates. Despite involving only a pre-trained teacher model, \textbf{DualDE} still encounters significant training costs due to the necessity of updating the teacher model during the second stage \cite{liu2023iterde}. \textbf{IteDE} \cite{liu2023iterde} introduces an iterative knowledge distillation approach aiming at progressively reducing model sizes to a predefined target. Throughout this process, multiple models of intermediate sizes are trained under the guidance of adjacent higher-dimensional models. Essentially, it remains within the realm of multiple-teacher strategies. These methods typically prioritize enhancing the performance of low-dimensional embeddings with limited regard for the associated training costs.

However, in the context of FKGE, the training procedure typically involves iterative communication between clients and a central server. Given constraints such as limited network bandwidth and client-side data usage, minimizing parameter transmission during FKGE training is crucial. From this perspective, the methods commonly employed in traditional KGE may not be directly applicable to FKGE due to their potential for significant communication costs. Hence, there is a need to devise more lightweight embedding compression techniques tailored specifically for FKGE.

\section{Preliminaries}
In this section, we explain the preliminary knowledge about KG, FKG, FKGE, \textbf{FedE} (the pioneering FKGE method) and the task of FKGE \textbf{C}ompression based on \textbf{K}nowledge \textbf{D}istillation.

A \textbf{K}nowledge \textbf{G}raph (\textbf{KG}) $\mathcal{G}$ with entity set $\mathcal{E}$ and relation set $\mathcal{R}$ is defined as $\mathcal{G}~ =~ \{(h,r,t)~|~h,t ~\in~ \mathcal{E}, ~ r ~\in ~ \mathcal{R}\}$, where each triplet $(h,r,t)$ represents the fact that the head entity $h$ has relation $r$ with the tail entity $t$. 

\textbf{F}ederated \textbf{K}nowledge \textbf{G}raph (\textbf{FKG}) is a set of KGs which are stored in multiple clients for the aim of data privacy. Formally, the definition of FKG is as follows:

\begin{definition}
\textbf{Federated Knowledge Graph: } Given $C$ clients, each client holds a knowledge graph $\mathcal{G}_c~ =~ \{\mathcal{E}_c, \mathcal{R}_c,\mathcal{T}_c ~|~ c\in [0, C]\}$, and the entity set in these $C$ knowledge graphs can overlap, but the relation sets are mutually exclusive. The set of these $C$ knowledge graphs is called Federated Knowledge Graph: $\mathcal{G}_F ~ = ~ \{\mathcal{G}_c\}^{C}_{c=1}$.
\end{definition}

\textbf{F}ederated \textbf{K}nowledge \textbf{G}raph \textbf{E}mbedding (\textbf{FKGE}) aims to let all KGs in FKGE collaboratively engage in embedding learning to achieve enhanced embeddings with improved performance, without exposing raw triples to each other. This collaborative process leverages the presence of shared entities among clients' KGs, allowing each client to contribute additional semantic information to others.

\textbf{FedE} is the first FKGE method which mainly includes two parts: client update and server update. We take the communication round $t$ as an example to explain them, considering the different rounds share the same process. For server updates, the server first collects the updated entity embeddings at round $t-1$ from each client. Subsequently, the server performs an average embedding aggregation for each entity. For client updates, each client initially receives the aggregated embeddings from the server and replaces their local entity embeddings with these updated versions. Following this, each client proceeds with local embedding learning using a KGE method. The objective of local embedding learning is scoring positive triples higher than negative ones based on the score function of the chosen KGE method. 

\textbf{FKGE} \textbf{C}ompression based on \textbf{K}nowledge \textbf{D}istillation (\textbf{FKGE-CKD}) aims at reducing the embedding dimension for a FKGE method using knowledge distillation techniques, and simultaneously avoiding decrease embedding performance significantly. In this paper, we further assume that the FKGE-CKD training process should follow the existing client and server communication mechanisms, and it should only intervene in the client local learning process. Other scenarios are reserved for future research. 
     
\section{Methodology}

In this section, we explain our proposed method \textbf{FedKD} in detail. 

\subsection{Overview of \textbf{FedKD}}

\textbf{FedKD}, as a supplementary component functioning only in client local training phrase, is orthogonal to existing FKGE models, which aims to decrease their embedding dimensions without significant performance decrease. Because all clients at each communication round share the same process, we take client $c$ as an example to explain it.

To reduce the embedding dimension of a FKGE model, it begins with pre-training  the FKGE mdoel with high embedding dimension. These obtained high-dimensional embeddings are collectively referred as the teacher model. Correspondingly, the FKGE model configured with the specified low embedding dimension is denoted as the student model. For the student model, during client $c$'s local training at round $t$, besides applying the original local training loss function of the FKGE method as the \textbf{hard label loss} $\mathcal{L}^H_c$,
the \textbf{FedKD} component also encourages the student model to mimic the score distributions of triples from the teacher, known as the \textbf{soft label loss } $\mathcal{L}^S_c$. Formally, the client $c$'s total local training loss at round $t$ is as follows:

\begin{equation}\label{tl}
    \mathcal{L}_c = \mathcal{L}^H_c + \lambda \frac{ \mathcal{L}^S_c}{\mathcal{L}^H_c + \mathcal{L}^S_c}
\end{equation}
where $\lambda \in \mathbb{R}^1$ is a constant; the term $\frac{\lambda}{\mathcal{L}^H_c + \mathcal{L}^S_c}$ is used to dynamically balance the hard and soft label loss, which is explained in the following parts in detail.  

In this work, we use the Kullback-Leibler (KL) divergence as the the soft label loss. Different from traditional knoweldge distillation with KL, we introduce the Adaptive Asymmetric Temperature Scaling mechanism into the KL-based soft label loss. It aims to mitigate the problem that the teacher can be too confident about the positive triple. When the teacher is too confident about the positive triple, the distribution of corresponding negative samples are less discriminative, which affects the performance of knowledge distillation \cite{li2022asymmetric}. Besides, the disparity in optimization objectives between hard label loss and soft label loss can inadvertently impact training efficiency \cite{liu2023iterde}. To deal with it, we begin local training by emphasizing the importance of the hard label loss, gradually shifting towards greater emphasis on the soft label loss as training advances.

We will explain \textbf{FedKD} in the following section in detail.

\subsection{Client Local Training with Knowledge Distillation}

During client local training at round $t$, for each triple $(h, r, t)$ of client $c$, we first generate a set of negative samples of size $n$: $N_c(h, r, t) = \{ (h, r, t'_i)| i\in [1, n];~ t'_i \neq t; ~  (h, r, t'_i) \notin \mathcal{T}_c\}$. Subsequently, we compute scores for the triple $(h, r, t)$ and its negative samples using the teacher and student models, respectively. For the triple $(h, r, t)$, its scores at student and teacher models are computed as follows:  (Similarly, this process applies to its negative samples.) 

\begin{equation}
    S^{(h,r,t)}_{stu} = S({(h,r,t)}; E_{stu}, R_{stu})
\end{equation}
\begin{equation}
    S^{(h,r,t)}_{tea} = S({(h,r,t)}; E_{tea}, R_{tea})
\end{equation}

\noindent where $S(\cdot)$ is the scoring function adopted in the original FKGE model; $E_{stu}$ and $R_{stu}$ ($E_{tea}$ and $R_{tea}$) are the entity and relation embeddings of student (teacher) model. 

The score of a triple indicates its likelihood of being predicted as positive. 
Therefore, we transfer knowledge from the teacher model to the student model by
aligning their score distributions for each positive triple and its corresponding negative samples. Formally, for the triple (h, r, t), we compute its soft label loss for the student model as follows:

\begin{equation}
    \mathcal{L}^{S}_{(h, r, t)} = KL(\mathcal{P}^{stu}_{(h, r, \{t, t'_i\})}, \mathcal{P}^{tea}_{(h, r, \{t, t'_i\})})
\end{equation}
where $\{t, t'_i\}$ ($i \in [1, n]$) denotes the tail entity set of triple $(h, r, t)$ and its negative samples; $KL(\cdot)$ denotes the Kullback-Leiber Divergence; $\mathcal{P}^{stu}_{(h, r, \{t, t'_i\})}$ and $\mathcal{P}^{tea}_{(h, r, \{t, t'_i\})}$ are the score distributions of the student and the teacher model about triple $(h,r,t)$ and its negative samples, respectively. For student model, $\mathcal{P}^{stu}_{(h, r, \{t, t'_i\})}$ is 
composed by the score probability $\mathcal{P}^{stu}_{(h, r, \hat{t})}$ ($\hat{t} \in \{t, t'_i\}$) of triple $(h,r,t)$ and its negative samples. $\mathcal{P}^{stu}_{(h, r, \hat{t})}$ is computed as follows:

\begin{equation}\label{sp}
    \mathcal{P}^{stu}_{(h, r, \hat{t})} = \frac{exp(S^{(h,r,\hat{t})}_{stu} / \tau)}{\sum_{\hat{t} \in \{t, t'_i\} } exp(S^{(h,r,\hat{t})}_{stu} / \tau)}
\end{equation}
where $\tau$ is the temperature for scaling scores of triples; $\mathcal{P}^{tea}_{(h, r, \{t, t'_i\})}$ can be computed in the similar way.

However, we note that \textbf{ATS} (Asymmetric Temperature Scaling) \cite{li2022asymmetric} highlights a potential issue in knowledge distillation when the teacher model exhibits over-confidence. Specifically, if the teacher assigns excessively high scores to the correct class and lower or less varied scores to incorrect classes, applying a uniform temperature to scale the logits of the neural network can lead to reduced class discriminability. This phenomenon can adversely impact the effectiveness of the knowledge distillation process. To address this issue, \textbf{ATS} proposes the use of two distinct temperature values: one for the correct class and another for the incorrect classes. Inspired by the insight from \textbf{ATS}, we introduce a novel mechanism called Adaptive Asymmetric Temperature Scaling. Unlike \textbf{ATS}, which employs two pre-defined temperature parameters shared across all input samples, our method adaptively learns the temperature for each positive sample while only keep the temperature for the negative samples pre-defined. This adaptive approach aims to tailor the temperature scaling to the specific characteristics of each positive sample, thereby enhancing the effectiveness of the knowledge distillation process.

Hence, we further improve the score distributions of the student and teacher model: $\mathcal{P}^{stu}_{(h, r, \{t, t'_i\})}$ and $\mathcal{P}^{tea}_{(h, r, \{t, t'_i\})}$. Specifically, we assign an adaptive temperature to each positive triple, determined by the teacher model's confidence in that triple. To quantify the teacher model's confidence in the positive triple $(h, r, t)$, we utilize its score probability (i.e., $\mathcal{P}^{tea}_{(h, r, \hat{t})}$ as shown in Eq. \ref{sp}. Here, we set the $\tau$ as $1$) in this paper. A higher score probability for a positive triple indicates greater confidence by the teacher model. We also think that other methods, such as using the entropy of the teacher's score distribution  over the set of positive and negative samples (i.e., $\mathcal{P}^{tea}_{(h, r, \{t, t'_i\})}$), could also be explored. We consider these alternative measures as potential avenues for future research.

\begin{table*}
  \caption{Comparison among different FKGE methods in terms of link prediction results. Bold denotes the best results. Particularly, we highlight our model not only in bold but also in red to indicate when it achieved the best results, for clarity.}\label{tab:main_re}
  \def\arraystretch{1.25}%
  \setlength{\tabcolsep}{1mm}{
  \begin{tabular*}{\textwidth}{@{\extracolsep\fill}lllccccccccccccccc}
    \toprule
    \multirow{2}{*}{Dataset} &  \multirow{2}{*}{Dim}  &  \multirow{2}{*}{Method}  && \multicolumn{4}{c}{TransE} & & \multicolumn{4}{c}{RotatE} & &\multicolumn{4}{c}{ComplEx} \\
    \cmidrule{5-8}
    \cmidrule{10-13}
    \cmidrule{15-18}
    &&& & MRR& Hits@1&Hits@5&Hits@10 & & MRR& Hits@1&Hits@5&Hits@10& & MRR& Hits@1&Hits@5&Hits@10\\
    \midrule
    \multirow{4}{*}{FB-R3}& \multirow{2}{*}{256}&\textbf{Local}&&0.3229   &0.1986   &0.4704   &0.5608   &&0.3409   &0.2249 &0.4778 &0.5643  && 0.3029  &0.2170  &0.4000  &0.4724 \\

    \cmidrule{3-18}
    
    &&\textbf{FedEH} &&\textbf{0.3612}   &\textbf{0.2316}   &\textbf{0.5158}   &\textbf{0.6070}   &&\textbf{0.3702}   &0.2405 &\textbf{0.5244} &0.6129  &&\textbf{0.3198}  &\textbf{0.2097}  &0.4472  &0.5346\\

    \cmidrule{2-18}
    
    &\multirow{2}{*}{128}&\textbf{FedEL} &&0.3517   &0.2243   &0.5028   &0.5938   &&0.3658   &\textbf{0.2410} &0.5129 &0.6037  &&0.2874  &0.1795  &0.4109  &0.4995\\
    \cmidrule{3-18}
    &&\textbf{FedEKD} &&0.3582   &0.2290   &0.5120   &0.6018   &&0.3685   &0.2378 &0.5239 & \textcolor{red}{\textbf{0.6137}}  &&0.3193  &0.2024  & \textcolor{red}{\textbf{0.4552}}  & \textcolor{red}{\textbf{0.5459}}\\

    \midrule

    \multirow{4}{*}{FB-R5}& \multirow{2}{*}{256}&\textbf{Local}&&0.3014   &0.1797   &0.4438   &0.5335   &&0.3193   &0.2098 &0.4472 &0.5335  &&0.3013  &0.2165  &0.3973  &0.4645 \\

    \cmidrule{3-18}
    
    &&\textbf{FedEH} &&\textbf{0.3626}   &\textbf{0.2317}   &\textbf{0.5195}   &\textbf{0.6102}   &&\textbf{0.3723}   &0.2405 &\textbf{0.5308} & \textbf{0.6184}  &&\textbf{0.3056}  &\textbf{0.1963}  &0.4316  &0.5205\\

     \cmidrule{2-18}
     
    &\multirow{2}{*}{128}&\textbf{FedEL} &&0.3490   &0.2236   &0.4973   &0.5892   &&0.3642   &0.2374 &0.5159 &0.6066  &&0.2764  &0.1674  &0.4011  &0.4905\\
    \cmidrule{3-18}
    &&\textbf{FedEKD} &&0.3599   &0.2307   &0.5155   &0.6037   &&0.3711   & \textcolor{red}{\textbf{0.2417}} &0.5260 &0.6165  && 0.3047  &0.1898  & \textcolor{red}{\textbf{0.4366}}  & \textcolor{red}{\textbf{0.5277}}\\

    \midrule

    \multirow{4}{*}{FB-R10}& \multirow{2}{*}{256}&\textbf{Local}&&0.2869   &0.1626   &0.4337   &0.5244   &&0.3038   &0.1991 &0.4255 &0.5095  &&\textbf{0.3002}  &\textbf{0.2120}  &0.4005  &0.4713 \\

    \cmidrule{3-18}
    
    &&\textbf{FedEH} &&\textbf{0.3517}   &0.2153   &\textbf{0.5167}   &\textbf{0.6104}   &&\textbf{0.3657}   &\textbf{0.2316} &0.5265 &0.6184  && 0.2986 &0.1813  &\textbf{0.4340}  &\textbf{0.5297}\\

    \cmidrule{2-18}
     
    &\multirow{2}{*}{128}&\textbf{FedEL} &&0.3431   &0.2119   &0.4999   &0.5935   &&0.3581   &0.2292 &0.5113 &0.6026  &&0.2453  &0.1589  &0.3362  &0.4181\\
    \cmidrule{3-18}
    &&\textbf{FedEKD} &&0.3501   & \textcolor{red}{\textbf{0.2154}}   &0.5109   &0.6025   &&0.3650   &0.2307 & \textcolor{red}{\textbf{0.5267}} & \textcolor{red}{\textbf{0.6188}}  &&0.2956  &0.1796  &0.4278  &0.5234\\

  \bottomrule
\end{tabular*}}

\end{table*}

Specifically, we calculate the temperature $\tau_{(h, r, t)}$ for each positive triple $(h, r, t)$ using a two-layer MLP with ReLU as the activation function and the hidden dimension as $32$. Formally,

\begin{equation}
    \tau_{(h, r, t)} = (\tau_{max} - \tau_{min}) \times \sigma(MLP(\mathcal{P}^{tea}_{(h, r, \hat{t})})) + \tau_{min}
\end{equation}
where $\tau_{max} $ and $\tau_{min}$ define the upper and lower bounds of the temperature range, respectively; $\sigma$ denotes the Sigmoid function.

Hence, the score distributions of both student and teacher model (i.e., $\mathcal{P}^{stu}_{(h, r, \{t, t'_i\})}$ and $\mathcal{P}^{tea}_{(h, r, \{t, t'_i\})}$) is further revised. Take student model as an example,  for the positive triple $(h, r, t)$, its score probability is adjusted as follows:

\begin{equation}
\mathcal{P}^{stu}_{(h, r, t)} = \frac{exp(S^{(h,r,t)}_{stu} / \tau_{(h, r, t)})}{\sum^n_{i=0} exp(S^{(h,r,t'_i)}_{stu}) / \tau) + exp(S^{(h,r,t)}_{stu} / \tau_{(h, r, t)})}
\end{equation}
For the negative triple $(h, r, t'_i)$, its score probability is adjusted as follows:

\begin{equation}
    \mathcal{P}^{stu}_{(h, r, t'_i)} = \frac{exp(S^{(h,r,t'_i)}_{stu} / \tau)}{\sum^n_{i=0} exp(S^{(h,r,t'_i)}_{stu}) / \tau) + exp(S^{(h,r,t)}_{stu} / \tau_{(h, r, t)})}
\end{equation}
where $\tau_{(h, r, t)}$ is the  adaptive temperature for positive triple (h, r, t) and $\tau$ is the pre-defined temperature for all negative triples.

The score distribution of teacher model $\mathcal{P}^{tea}_{(h, r, \{t, t'_i\})}$ is revised in the similar way.

The divergence in optimization goals between hard label loss and soft label loss may inadvertently affect the efficiency of training \cite{liu2023iterde}. We opt to prioritize the hard label loss during the initial stages of local training, gradually augmenting the weight of the soft label loss to emphasize its influence during the later stages of local training. In our approach, we scale the soft label loss $\mathcal{L}^S_c$ with the coefficient $\frac{\lambda}{\mathcal{L}^H_c + \mathcal{L}^S_c}$, as the Eq. \ref{tl} shows. As training progresses and both hard and soft label losses diminish towards model convergence, this coefficient increases. Consequently, the influence of the soft label loss intensifies over the course of training.  

\section{Experiment}

In this section, we apply the proposed \textbf{FedKD} to the seminal FKGE model \textbf{FedE} and evaluate the effectiveness of embedding compression.

\subsection{Experiment Setup}

\subsubsection{Dataset}

We perform experiments using three widely used datasets for FKGE: FB-R3, FB-R5, and FB-R10. These datasets are derived by evenly partitioning relations from the FB15k-237 dataset and distributing associated triples across three, five, and ten clients respectively. Each dataset maintains the same division ratio for training, validation, and testing sets: 0.8/0.1/0.1.

\subsubsection{Baselines}

We apply the proposed \textbf{FedKD} component on the initial FKGE model \textbf{FedE}. To leverage the benefits of personalized modeling, we enhance \textbf{FedE} by adopting local embeddings to evaluate the performance on validation and testing sets instead of using global embeddings throughout the training process. The low-dimensional \textbf{FedE} integrated with the \textbf{FedKD} component is referred to as the student model, denoted as \textbf{FedEKD} for simplicity. The high-dimensional \textbf{FedE} serves as the teacher model and acts as a baseline for comparison. The model is denoted as \textbf{FedEH} for convenience. Additionally, we use \textbf{FedE} with the same dimension as \textbf{FedEKD} as another baseline, labeled as \textbf{FedEL}. Furthermore, we include a model (i.e., \textbf{Local}) trained only on local triples, without federated learning, but at the same high dimension as the teacher model, as an additional baseline for our experiments. During the clients' local training phase, we follow the standard practice of utilizing three representative KGE methods from previous FKGE studies: TransE \cite{transe}, RotatE \cite{rotate}, and ComplEx \cite{complex}. 

\subsubsection{Evaluation Metrics}

We evaluate the model performance by link prediction task which predicts the the tail (head) entity when given the head (tail) entity and relation. The link prediction evaluation metric are Mean Reciprocal Rank (MRR) and Hits@N (N is 1, 5 and 10). The overall metric value across clients is computed by weighted average, where the weights correspond to the proportions of the triple sizes across the clients. 

\subsubsection{Implementation Details}\label{id}

We assume full participation of all clients in each communication round. The parameters in the experiments are as follows. For client training, the batch size is set to 512 and the number of local epochs is 3. The high and low embedding dimension are set as 256 and 128, respectively. The initialization parameter $\epsilon$ is set as $2$ for both high and low embedding dimension, while another initialization parameter $\gamma$ is set as $8$ and $6$ for high and low embedding dimension, respectively. During training, we employ an early stopping strategy with a patience threshold of $3$. For the \textbf{Local} model, performance evaluation occurs every $10$ epochs, while for all other models, evaluation takes place every $5$ epochs. The optimizer Adam \cite{adam} is adopted and the learning rate is set as $0.0001$. The papameter $\lambda$ is set as $3$. The parameter $\tau_{min}$ and $\tau$ is set as $1$ for all cases. For dataset FB-R3, the choice of $\tau_{max}$ is $2$ for TransE and RotatE, and $1.5$ for ComplEx. For dataset FB-R5, $\tau_{max}$ is set to $1.5$ for RotatE and ComplEx. In dataset FB-R10, $\tau_{max}$ is $1.5$ for TransE. In all other cases, we set the default value of $\tau_{max}$ to $10$.

\subsection{Main Experiment}
In this section, we present a quantitative assessment of the effectiveness of the \textbf{FedKD} component. This evaluation is conducted by comparing the performance metrics of \textbf{FedEKD}—a variant derived by integrating \textbf{FedKD} into \textbf{FedE}—against established baseline models. The experimental results across three distinct datasets are detailed in Table \ref{tab:main_re}.

Analysis of the data presented in Table \ref{tab:main_re} reveals a marked decline in performance for the \textbf{FedEL} model compared to \textbf{FedEH}. This decline is particularly pronounced when utilizing ComplEx as the KGE method. In particular, for the FB-R3, FB-R5, and FB-R10 datasets, the \textbf{FedEL} model shows a decrease in MRR by 10.13\%, 9.55\%, and 17.85\%, respectively. Similar downward trends are also observed in the Hits@1, Hits@5, and Hits@10 metrics. These findings indicate that training low-dimensional FKGE directly leads to suboptimal performance outcomes. 

When comparing \textbf{FedEL} and \textbf{FedEKD}, both operating in a 128-dimensional space, \textbf{FedEKD} shows substantial performance improvement. Furthermore, \textbf{FedEKD} achieves performance levels that approximate those of the teacher model \textbf{FedEH} with higer dimension(256).
Specifically, when TransE is employed as the KGE method, \textbf{FedEKD} achieves a relative increase in MRR of 1.85\%, 3.12\%, and 2.04\% on the FB-R3, FB-R5, and FB-R10 datasets, respectively. \textbf{FedEKD} attains performance levels of 99.17\%, 99.26\%, and 99.55\% relative to \textbf{FedEH}.

When Rotate is used as the KGE method, compared with \textbf{FedEL}, \textbf{FedEKD} achieves the relative increases in MRR on these datasets are 0.74\%, 1.89\%, and 1.93\%. It finally attains MRR performance levels of 99.54\%, 99.68\%, and 99.81\% relative to \textbf{FedEH}.
Notably, with ComplEx as the KGE method, \textbf{FedEKD}, compared with \textbf{FedEL} shows a more pronounced improvement in MRR, with increases of 11.10\%, 10.24\%, and 20.51\% on the FB-R3, FB-R5, and FB-R10 datasets, respectively. It achieves MRR levels of 99.84\%, 99.74\%, and 99.00\% relative to \textbf{FedEH}. A significant upward trend of \textbf{FedEKD} compared with \textbf{FedEL} is also observed across the metrics Hits@1, Hits@5, and Hits@10. Unexpectedly, \textbf{FedEKD} even outperforms \textbf{FedEH} in certain cases, in terms of Hits@1, Hits@5, and Hits@10 metrics. These results collectively demonstrate that the \textbf{FedKD} component facilitates a reduction in FKGE embedding dimensions from 256 to 128 with minimal to negligible performance degradation.

\begin{table}
  \caption{Ablation study about the Adaptive Asymmetric Temperature Scaling mechanism on dataset FB-R10. The model \textbf{FedEKD} incorporates the mechanism while \textbf{FedEKD*} does not. Bold denotes the best results.}\label{tab:as}
  \def\arraystretch{1.25}%
  \setlength{\tabcolsep}{0.8mm}{
  \begin{tabular}{lllccccc}
    \toprule
    \multirow{2}{*}{KGE} &  \multirow{2}{*}{Dim}  &  \multirow{2}{*}{Method}  && \multicolumn{4}{c}{Metric}\\
    \cmidrule{5-8}
    &&& & MRR& Hits@1&Hits@5&Hits@10\\
    \midrule
    \multirow{2}{*}{TransE} &\multirow{2}{*}{128}& \textbf{FedEKD*} &&0.3490   &0.2134   &0.5102   &\textbf{0.6027} \\
    \cmidrule{3-8}
    &&\textbf{FedEKD} &&\textbf{0.3501}   & \textbf{0.2154}   &\textbf{0.5109}   &0.6025 \\

    \midrule
    
    \multirow{2}{*}{RotatE}& \multirow{2}{*}{128}&\textbf{FedEKD*} &&0.3635   &\textbf{0.2312} &0.5221 &0.6134\\
    \cmidrule{3-8}
    &&\textbf{FedEKD} &&\textbf{0.3650}   &0.2307 & \textbf{0.5267} & \textbf{0.6188} \\
    \midrule
    \multirow{2}{*}{ComplEx}&\multirow{2}{*}{128}&\textbf{FedEKD*} &&0.2940  &0.1788  &0.4248  &0.5198\\
    \cmidrule{3-8}
    &&\textbf{FedEKD} &&\textbf{0.2956}  &\textbf{0.1796}  &\textbf{0.4278}  &\textbf{0.5234}\\

  \bottomrule
\end{tabular}}

\end{table}

\subsection{Ablation Study}

In this section, we investigate whether the Adaptive Asymmetric Temperature Scaling mechanism can improve the performance. We first remove the mechanism from the component $\textbf{FedKD}$ and then apply the remains of $\textbf{FedKD}$ to $\textbf{FedE}$. The derived model is denoted \textbf{FedEKD*} for clarity. We follow the same setting for the other parameters as described in \ref{id} and conduct experiments to compare the performance of $\textbf{FedEKD}$ and \textbf{FedEKD*}. The result is shown in the Table \ref{tab:as}. 

The data presented in the table indicates that \textbf{FedEKD} generally outperforms \textbf{FedEKD*} across all evaluated metrics, except for Hits@1 with the RotatE KGE method and Hits@10 with the TransE KGE method. Moreover, compared to TransE, both RotatE and ComplEx exhibit more pronounced improvements overall. Specifically, the increase in MRR is 0.11\% for TransE, 0.15\% for RotatE, and 0.16\% for ComplEx. For RotatE and ComplEx, the enhancements in metric Hits@5 and Hits@10 are more notable than the metric MRR. RotatE shows an increase of 0.46\% in Hits@5 and 0.54\% in Hits@10, while ComplEx demonstrates increases of 0.30\% and 0.36\%, respectively. These improvements underscore the efficacy of the Adaptive Asymmetric Temperature Scaling mechanism.

\section{Conclusion}

FKGE enables collaborative learning of entity and relation embeddings across distributed knowledge graphs while ensuring data privacy. Higher-dimensional embeddings are typically favored in FKGE models for enhanced performance, yet they introduce challenges like resource-intensive storage and slower inference. Unlike traditional methods, FKGE involves iterative client-server communication, requiring efficient data transfer. Compression techniques for traditional KG embeddings may not suit FKGE due to their reliance on multiple model training and potential communication costs. This study introduces \textbf{FedKD}, a novel component tailored for FKGE, utilizing Knowledge Distillation to enhance training on client devices. \textbf{FedKD} adjusts temperature parameters dynamically for positive triples to refine score distributions, mitigating issues of teacher model over-confidence. Additionally, adaptive adjustment of KD loss weighting optimizes the training process. Experimental results indicate that \textbf{FedKD} achieves a reduction in embedding dimensions by half with minimal to negligible degradation in performance.

\newpage
\bibliographystyle{ACM-Reference-Format}
\bibliography{sample-base}

\end{document}